\documentclass[10pt,twocolumn,letterpaper]{article}

\pdfoutput=1

\usepackage{cvpr}
\usepackage{times}
\usepackage{epsfig}
\usepackage{graphicx}
\usepackage{amsmath}
\usepackage{amssymb}
\usepackage{algorithm}
\usepackage{algorithmicx}
\usepackage{multirow}
\usepackage{stfloats}
\usepackage[noend]{algpseudocode}
\usepackage{float}
\usepackage{setspace}

\usepackage[breaklinks=true,bookmarks=false]{hyperref}

\cvprfinalcopy % *** Uncomment this line for the final submission

\def\cvprPaperID{****} % *** Enter the CVPR Paper ID here

\begin{document}

%%%%%%%%% TITLE
\title{Weakly-supervised Learning of Mid-level Features for Pedestrian Attribute Recognition and Localization}

\author{Kai Yu\\
Beihang Univeristy\\
Beijing, China\\
{\tt\small kyu\_115s@buaa.edu.cn}
\and
Biao Leng\\
Beihang Univeristy\\
Beijing, China\\
{\tt\small lengbiao@buaa.edu.cn}
\and
Zhang Zhang\\
Chinese Academy of Sciences\\
Beijing, China\\
{\tt\small zzhang@nlpr.ia.ac.cn}
\and
Dangwei Li\\
Chinese Academy of Sciences\\
Beijing, China\\
{\tt\small dangwei.li@nlpr.ia.ac.cn}
\and
Kaiqi Huang\\
Chinese Academy of Sciences\\
Beijing, China\\
{\tt\small kqhuang@nlpr.ia.ac.cn}
}

\maketitle
%\thispagestyle{empty}

%%%%%%%%% ABSTRACT
\begin{abstract}
State-of-the-art methods treat pedestrian attribute recognition as a multi-label image classification problem. The location information of person attributes is usually eliminated or simply encoded in the rigid splitting of whole body in previous work. In this paper, we formulate the task in a weakly-supervised attribute localization framework. Based on GoogLeNet, firstly, a set of mid-level attribute features are discovered by novelly designed detection layers, where a max-pooling based weakly-supervised object detection technique is used to train these layers with only image-level labels without the need of bounding box annotations of pedestrian attributes. Secondly, attribute labels are predicted by regression of the detection response magnitudes. Finally, the locations and rough shapes of pedestrian attributes can be inferred by performing clustering on a fusion of activation maps of the detection layers, where the fusion weights are estimated as the correlation strengths between each attribute and its relevant mid-level features. Extensive experiments are performed on the two currently largest pedestrian attribute datasets, i.e. the PETA dataset and the RAP dataset. Results show that the proposed method has achieved competitive performance on attribute recognition, compared to other state-of-the-art methods. Moreover, the results of attribute localization are visualized to understand the characteristics of the proposed method.
\end{abstract}

%%%%%%%%% BODY TEXT
\section{Introduction}

The recognition of pedestrian attributes, such as gender, glasses and wearing styles, has become a hot research topic in recent years, due to its great application potentials in video surveillance systems, e.g. pedestrian re-identification where attributes can serve as mid-level representations of a pedestrian to improve the accuracy of ReID effectively \cite{layne2012person}, and pedestrian retrieval where the queried attributes can be used to filter out the interesting targets from a large amount of videos efficiently \cite{vaquero2009attribute, feris2014attribute}.

Pedestrian attribute recognition in surveillance scene is also a challenging problem due to the low resolution of pedestrian samples cropped from far-range surveillance scenes, the large pose variations arisen from different angles of view, occlusions from environmental objects, etc. Recently, convolutional neural network (CNN) has been applied for pedestrian attribute recognition \cite{zhu2015multi, sudowe2015person, li2015multi}, where high classification accuracies have been reported. In these work, pedestrian samples cropped out from scenes are fed into an end-to-end CNN classifier outputing multiple pedestrian attribute labels. Nevertheless, to enhance the performance of attribute recognition, there are still a number of problems worthy of further studies. Firstly, some fine-scale attributes such as glass wearing are hard to recognize due to the small size of positive samples. Secondly, some appearance features of these fine-scale attributes themselves may be easily neglected during the several alternations of convolution and max-pooling operations, so the final prediction layers of deep models cannot encode all the detailed features of fine-scale attributes for correct attribute predictions. Thirdly, the locations of some attributes can vary significantly in the cropped pedestrian sample. For example, when saying a pedestrian is carrying a bag, the vertical location of the bag may range from his arms to his knees, which introduces difficulty into training of the traditional CNNs. Finally, the pedestrian himself may appear at unusual regions of the cropped image samples, while some previous methods are developed under the assumption that the pedestrian appears in the middle and occupies most of the area of the image, so as to utilize some predefined spatial distributions of attributes for better performance. This problem can be serious when the pedestrian samples are cropped automatically by some pedestrian detection algorithms. Besides, the previous work shows little effect in locating attributes, thus the location information of attributes cannot be utilized by follow-up algorithms.

Considering the above difficulties, we formulate pedestrian attribute recognition in an attribute localization framework, where a Weakly-supervised Pedestrian Attribute Localization Network (WPAL-network) is proposed to infer the attribute labels from the detection results of a set of mid-level attribute-relevant features, instead of the direct classification from whole pedestrian samples. The motivation lies in that solid and abstract attributes are all relevant to some special kinds of mid-level semantic features which may be obtained by deep learning at some high level layers \cite{le2013building}. For example, whether a pedestrian is carrying a bag can be directly determined by the detection of appearance feature of a bag being carried, and female gender can be easily inferred if long hair or a miniskirt is detected. Recognizing attributes by flexibly detecting these features in an image without resizing and warping can eliminate the problem that pedestrians have different statures and may appear in unusual regions in the sample image.

Since it is high-cost to label the exact locations of multiple attributes across a large dataset, and some attributes like glass wearing have ambiguous bounding box definition, the powerful fully-supervised object detection methods are not applicable. Therefore we only use image-level attribute labels to conduct weakly-supervised learning to discover mid-level semantic features with a set of weakly-supervised detection layers. These layers are similar to the network structure proposed in \cite{oquab2015object}, which focuses on general object detection with only image-level absence/presence labels. In this paper, we modified the structure to adapt to the pedestrian attribute localization problem. One difference is that we use Flexible Spatial Pyramid Pooling (FSPP) instead of the original global max-pooling to add spatial constraint to some attributes like hats. Another is that the structure lays in the middle stage of the network but not the top, making correlation between detector and target class not bound at first but free to be learnt during training.

With the trained WPAL-network, we can locate an attribute according to the responses of the detectors of discovered mid-level attribute features. The correlation strength between attributes and the mid-level features is firstly statistically estimated over the training set with the trained network. Then, a rough shape of attribute is estimated by superposing activation maps of the mid-level detectors with weight as the correlation strength. Finally, the location of attribute is predicted as the centroid of activation cluster.

To demonstrate the effectiveness of the proposed network, extensive experiments are performed on the two large-scale pedestrian attribute datasets, i.e., PETA and RAP. Compared to the state-of-the-art methods, the WPAL-network can achieve competitive performance. And the results of attribute localizations can be visualized to further explain the characteristics of the proposed method.

The contributions of this work are concluded as follows:
\begin{itemize}
\item[-] We introduce weakly-supervised object detection technique into solving the pedestrian attribute recognition task, achieving state-of-the-art accuracy.
\item[-] The proposed method can not only predict existence labels of attributes but also locate the attributes, so as to provide location information for further applications.
\end{itemize}

The remainder of this paper is structured as follows: In Section~\ref{sec:rel}, we review previous work related to the method proposed in this paper in different aspects. In Section~\ref{sec:net}, the WPAL-network is illustrated in details. In Section~\ref{sec:loc}, we describe the method of attribute localizing using the WPAL-network. In Section~\ref{sec:exp}, we show some results of experiments on attribute recognition and attribute localizing.

%-------------------------------------------------------------------------
\section{Related Work}\label{sec:rel}

\begin{figure*}[!tbp]
  \centering
  \includegraphics[width=0.95\textwidth]{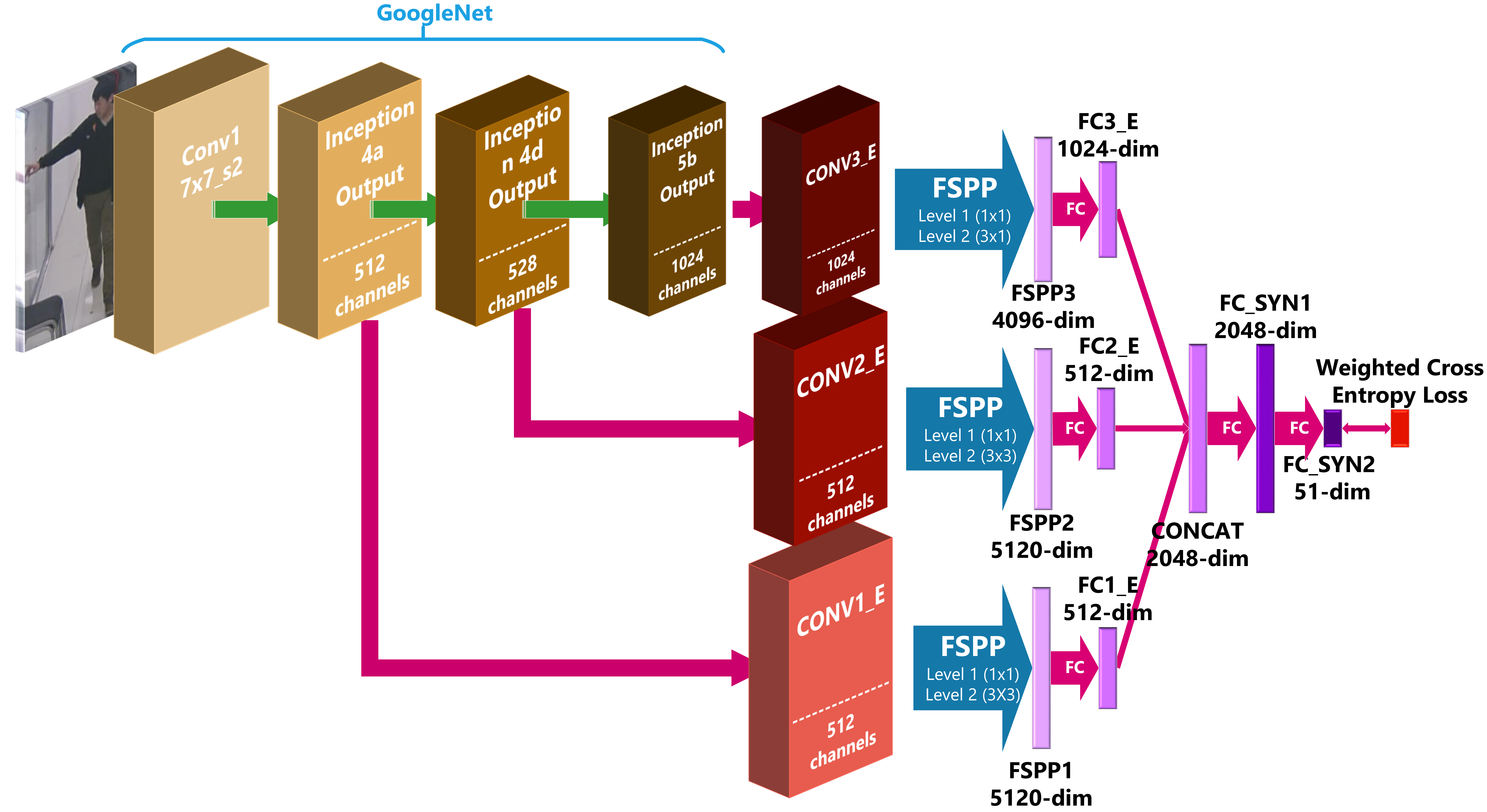}\\
  \caption{The overall architecture of the WPAL-network. The input image goes through the trunk convolution layers derived from GoogLeNet. Then three data flows branch out from different level. After FSPP layers, the resulting vectors are concatenated for final attribute prediction. Best viewed in color.}
  \label{fig:Architecture}
\end{figure*}

In this section, we firstly review the developments on pedestrian attribution recognition. Then, some related work on weakly-supervised object detection is introduced, which inspires us to develop the new solution on attribute localization.

\subsection{Pedestrian Attribute Recognition}

Early work \cite{sharma2011learning, sharma2013expanded, bourdev2011describing} on human attribute recognition usually treat attributes as independent labels and train classifiers for each attribute independently. Deep learning models used in some later work enabled researchers to mine the relationship between attributes. Patrick et al. proposed the ACN model in \cite{sudowe2015person} to jointly learn all the attributes in a single model, and showed that parameter sharing can improve recognition accuracy over independently trained models. This routine is also adopted in the DeepMAR model proposed in \cite{li2015multi} and the WPAL-network in this work.

It is yet another a popular idea to make use of part information to help improving the attribute recognition accuracy. In \cite{bourdev2014pose}, part models like DPM and poselets are used for aligning input patches for CNNs. Gaurav et al. propose an expanded parts model in \cite{sharma2013expanded} to learn a collection of part templates which can score an image partially with most discriminative regions for classification. The MLCNN in \cite{zhu2015multi} divides a human body into 15 parts and train CNN models for each of them, then choose part of the models to contribute to the recognition of an attribute, according to the spatial constraint prior of it. The DeepMAR* model described in \cite{li2016richly} takes three block images as input in addition to the whole body image, which correspond to the head-shoulder part, upper body and lower body of a pedestrian respectively. The idea of dividing the image into parts is adopted in the design of the WPAL-network, which drives us to make use of flexible spatial pyramid pooling layers to help locating mid-level features of some attributes in only local patches rather than the whole image.

\subsection{Weakly-supervised Object Detection}

To avoid the high-cost of labeling bounding boxes of objects, researchers proposed various weakly-supervised learning approaches for object detection and localization. In \cite{pandey2011scene}, Pandey et al. demonstrate capability of SVM and deformable part models on weakly-supervised object detection. In \cite{wang2014weakly}, Wang et al. proposed unsupervised latent category learning, which can discover latent information in backgrounds to help object localization in cluttered backgrounds. Cinbins et al. proposed in \cite{cinbis2014multi} a multi-fold multiple-instance learning procedure featuring prevention of weakly-supervised training from prematurely locking onto erroneous object locations.

In \cite{oquab2015object}, the proposed network has convolution layers followed by a global max-pooling layer. Each channel of the global max-pooling layer is viewed as a detector for a certain class of object. It is assumed that the positions of max value point in the feature map correspond to the locations where the objects of the target class exist in. However, this method cannot be directly applied to our attribute localization task. Firstly, different from objects, some attributes are abstract concepts, such as gender, orientation and age, which do not correspond to certain regions. Secondly, some attributes such as hat wearing or shoe style are expected to appear within a certain partition in a pedestrian sample, which can be used to improve the localization of those attributes. Thus, to better fit the task of attribute localization, we embed this structure in the middle stage of the network to discover mid-level features relevant to attributes rather than attributes themselves, and propose to use FSPP layers instead of a single global max-pooling layer to help constraining location of certain attributes.

%-------------------------------------------------------------------------
\section{Weakly-supervised Pedestrian Attribute Localization Network}\label{sec:net}

In this section, we describe the proposed WPAL-Network. The overall architecture is firstly illustrated and then detailed implementation is discussed.

\subsection{Network Architecture}

The framework of the WPAL-network is illustrated in Figure~\ref{fig:Architecture}. The trunk convolution layers are derived from the GoogLeNet model \cite{szegedy2015going} pretrained on ImageNet, provided by Caffe \cite{jia2014caffe}. In the original GoogleNet, the "inception4a/output", "inception4d/output" and "inception5b/output" layers are connected to some branch layers respectively. In the WPAL-network, each branch is replaced by a convolution layer then followed by flexible spatial pyramid pooling (FSPP) layers.

\begin{figure}[!tbp]
  \centering
  \includegraphics[width=0.48\textwidth]{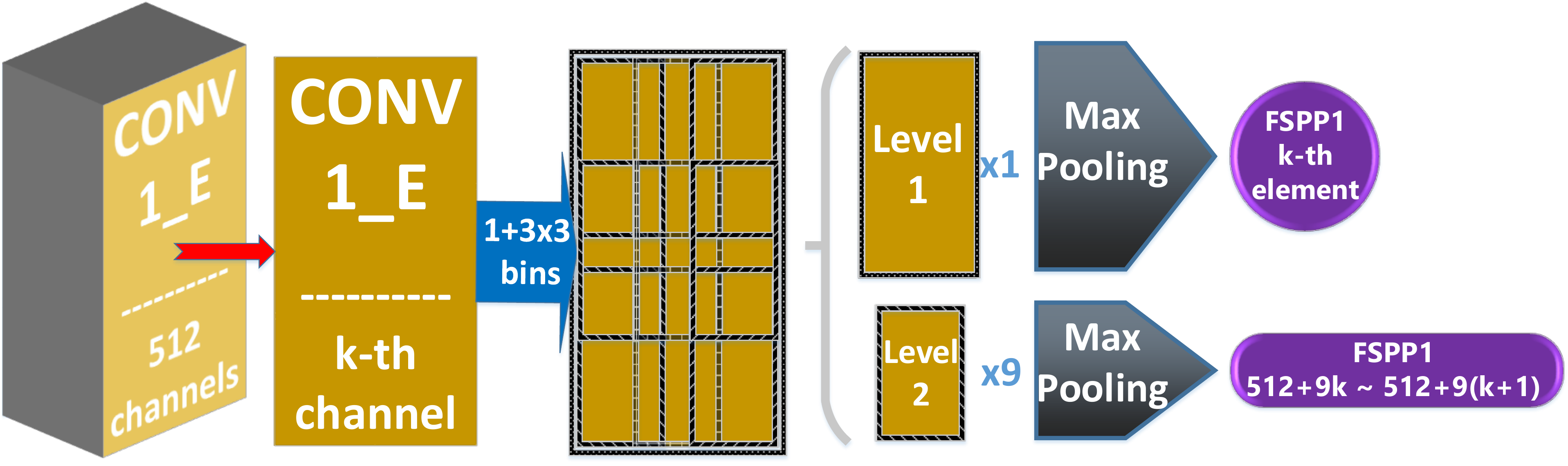}\\
  \caption{This picture illustrates the structure of a two-level FSPP layer. For a feature map in the bottom layer, on the first level, the bin covers the whole image; on the second level, several overlapping small bins together covers the image. For each bin, max-pooling is performed. The results are concatenated into a large vector of the whole FSPP layer. Best viewed in color.}
  \label{fig:FSPP}
\end{figure}

The FSPP layers play the role of the global max-pooling layers in \cite{oquab2015object} for the discovery of attribute-relevant features. Its mechanism is shown in Figure~\ref{fig:FSPP}, where the SPP layers in \cite{he2014spatial} are extended by allowing bins to overlap and the number of bins at each pyramid level to be changeable. At the first pyramid level, there is only one bin for each FSPP layer, which outputs the maximal response of each convolution channel over the full image. At the second level, the three FSPP layers are divided into $3 \times 1$, $3 \times 3$ and also $3 \times 3$ bins respectively, where max-pooling is performed in each bin. To avoid a high dimension output vector, we limit the height of pyramid to 2. For one FSPP layer, each convolution channel will produce a small vector with the dimension of the total number of bins at all the pyramid levels. Finally, these small vectors of all FSPP layers and all channels are concatenated into a large vector, which is further regressed into a 51-dim vector (35-dim for PETA dataset), corresponding to the attribute labels to be predicted.

To understand this architecture, we first discuss the function of the max-pooling operations in the FSPP layers. For each region of a convolution channel corresponding to a bin in a FSPP layer, the global max-pooling output indicates the possibility of certain mid-level feature exists or not, where the position of the maximal response is also expected to be the location of the mid-level feature.  The bins thus can be viewed as local detectors of the mid-level feature, except the bin on the first level which is a global detector. With the following fully-connected layers, the vector composed of the existence possibility values of mid-level features is regressed to form the attribute vector, where the correlations between mid-level features and attributes are encoded in the learnt weighting coefficients.  The training procedure therefore has two tasks. The first one is to learn the correlation between attributes and randomly initialized mid-level detectors. The second one is to adapt the target mid-level features of detectors to fit the correlated attributes. These two tasks can be solved simultaneously throughout the whole training process. With the learnt correlation, the detection results of mid-level features can also be used for attribute localization (see Section~\ref{sec:loc}).

If the network performs only global max-pooling over the full image rather than using the FSPP layers, it cannot get satisfying results of attribute localization, although it works well in attributes classification (shown in Section~\ref{sec:AccExp}, named as WPAL-GoogleNet-GMP). That is because the single global max-pooling usually leads to multiple activation clusters in different locations corresponding to multiple attribute-relevant mid-level features, which makes it difficult to determine which activation point can be used to infer the location of attributes. The FSPP layers are proposed to address this problem, based on the observation that mid-level features contributing to certain attributes have a local spatial distribution. For example, the features relevant to the attribute of hat wearing mostly appear in the upper part of a pedestrian. Thus, we adopt the multiple local max pooling operations within the bins defined by the second pyramided level in the FSPP layers as well as the global max-pooling at the first pyramid level independently. Note that we do not manually set which bin to favor for an attribute. This is also to be learnt during the training process. For example, for a detector which is expected to detect mid-level features that contribute to upper part attributes, the weights of connections from the upper bins will increase since these bins are activated more accordantly with the positive attribute labels, while the weights of connections from the lower bins get suppressed by weight-decay. For mid-level features that may equally appear in any part of the image, the bin on the first pyramid level of the FSPP layer, which is equivalent to global max-pooling, is then favored.

In this work, the shape and size of input image is not fixed, because the feature maps with variant sizes will be turn into vectors with fixed dimensionality same as the number of convolution channels after the max-pooling operations in the FSPP layers. This means the WPAL-network can process images of arbitrary resolutions without warping or transforming in preprocessing. Thus, the original shape information of the pedestrian body and other accessories can be preserved.

\subsection{Multi-level Learning}

The trunk layers are pretrained on the ImageNet dataset, in order to learn general features for a wide range of objects. As we know, the features' abstraction level will increase along with the convolution levels. However, the pedestrian attributes locate at different scale and abstraction level. For example, the orientation of whole body is at a higher level than that of attribute of "wearing glasses or not". Therefore, we need utilize information at different scale and abstraction levels for multiple attributes recognition. Here, the relatively general features learnt by three selected trunk convolution layers, i.e. Inception4a/output, Inception4d/output and Inception5b/output, are selected to be transformed by the CONVx\_E layers to fit the attribute features. Note that instead of explicitly specifying the learning levels of certain attributes, the decision is learnt by training the fully-connected layers, similar to the learning of attribute-detector correlations.

\subsection{Loss Function For Unbalanced Training Data}

For multi-label recognition tasks, usually the cross entropy loss function is adopted:

\begin{equation}
    \label{eq:CrossEntropyLoss}
    Loss_{ce} = \sum_{i=1}^{L} p_i \cdot log(\hat{p_i}) + (1 - p_i) \cdot log(1 - \hat{p_i})
\end{equation}

However, in the pedestrian attribute datasets (e.g., the PETA dataset \cite{li2016richly} and the RAP dataset \cite{deng2014pedestrian}), the distributions of positive and negative labels in most attribute classes are usually imbalanced. Many attributes, such as "wearing a V-neck or not", are seldom labeled positive in the training data. Using the objective function in Equation~\ref{eq:CrossEntropyLoss} may cause these attributes to be constantly predicted as negative. To address this problem, we introduce a weighted cross entropy loss function as follows:

\begin{equation}
    \label{eq:WeightedCrossEntropyLoss}
    \begin{split}
        Loss_{wce} = & \sum_{i=1}^{L} \frac{1}{2w_i} \cdot p_i \cdot log(\hat{p_i}) \\
                     & + \frac{1}{2(1 - w_i)}(1-p_i) \cdot log(1 - \hat{p_i})
    \end{split}
\end{equation}

where $L$ is the number of attributes; $p$ is the ground-truth attribute vector, and $\hat{p}$ is the predicted attribute vector; $w$ is a weight vector indicating the proportion of positive labels over all attribute categories in the training set.

\section{Attribute Localization \& Shape Estimation}\label{sec:loc}

\begin{algorithm}[!tbp]
    \caption{Estimate bin-attribute relationship strength.}
    \label{alg:EstRel}
    \renewcommand{\algorithmicrequire}{\textbf{Input:}}
    \renewcommand{\algorithmicensure}{\textbf{Output:}}
    \begin{algorithmic}[1]
        \Require
            \Statex $N$: Number of training samples
            \Statex $M$: Number of detectors
            \Statex $NB$: Number of bins in each detector
            \Statex $Score$: Matrix of bin output scores on training samples
            \Statex $Label$: Label vector of attribute $a$ of training samples
        \Ensure
            \Statex $PAve$: Average score on positive samples of $a$
            \Statex $NAve$: Average score on negative samples of $a$
            \Statex $RS$: Relationship strength array.
        \Procedure{Estimate}{$N, M, NB, Score, Label$}
            \For{$d := 1$ to $M$; $k := 1$ to $NB_d$}
                \begin{spacing}{0.2}\end{spacing}
                \State $PAve_{d_k} \gets \frac{\sum_{i=1}^{N} Score_{d_k,i} \cdot Label_i}{\sum_{i=1}^{N} Label_i}$
                \begin{spacing}{0.2}\end{spacing}
                \State $NAve_{d_k} \gets \frac{\sum_{i=1}^{N} Score_{d_k,i} \cdot (1 - Label_i)}{\sum_{i=1}^{N} (1 - Label_i)}$
                \begin{spacing}{0.2}\end{spacing}
                \State $RS_{d_k} \gets \frac{pos_{d_k}}{neg_{d_k}}$
                \begin{spacing}{0.2}\end{spacing}
            \EndFor
            \State \Return{$PAve,NAve,RS$}
        \EndProcedure
    \end{algorithmic}
\end{algorithm}

To locate an attribute $a$, we first determine the strength of correlation between the attribute and the bins of mid-level feature detectors. For each detector bin, the correlation strength is calculated as a ratio of the average score on positive samples to the average score on negative samples. This process can be formulated as Algorithm~\ref{alg:EstRel}.

Then, an existence possibility map $D$ of the attribute can be estimated by superposing the weighted activation maps masked by Gaussian filter, where the weights are the normalized correlation strengths, as shown in Algorithm~\ref{alg:Shape}. The extent of active region in $D$ indicates the rough shape of the attribute.

To locate the coordinator of the attribute, we perform clustering on the $D$, where the coordinators of the pixels that value greater than average value are collected for a weighted K-means clustering procedure. The pixel values also sever as the weights of each coordinate samples. Finally, the greatest several clusters are chosen as the candidates of attribute locations. The number of candidate clusters depends on the type of the attribute (e.g. 2 candidate clusters for shoes and 1 candidate cluster for hat).

\begin{algorithm}[!tbp]
    \caption{Estimate rough shape of an attribute}
    \label{alg:Shape}
    \renewcommand{\algorithmicrequire}{\textbf{Input:}}
    \renewcommand{\algorithmicensure}{\textbf{Output:}}
    \begin{algorithmic}[1]
        \Require
            \Statex $W,H$: Size of image and feature maps before detectors
            \Statex $X,Y$: Coordinators of detected mid-level features
            \Statex $A$: Activation maps of the detectors
        \Ensure
            \Statex $D$: Possibility distribution map of the attribute
        \Procedure{Localize}{$W,H,X,Y$}
            \Procedure{NormScore}{$\vec{s}, \hat{p_a}$}
                \For{$d := 1$ to $M$; $k := 1$ to $NB_d$}
                    \State $\vec{ns}_{d_k} \gets \vec{s}_{d_k} / \hat{p_a} ? PAve_{d_k} : NAve_{d_k}$
                \EndFor
                \State \Return{$\vec{ns}$}
            \EndProcedure
            \Procedure{Gauss}{$shape,c,var$}
                \For{$i := 1$ to $shape_h$; $j := 1$ to $shape_w$}
                    \State $mask_{i,j} \gets exp(-\frac{(i-c_y)^2 + (j-c_x)^2}{2 \cdot var})$
                \EndFor
                \State \Return mask
            \EndProcedure
            \State $norm\_score = NormScore(score, pred)$
            \State $D \gets EmptyMap(H,W)$
            \For{$d := 1$ to $M$; $k := 1$ to $NB_d$}
                \begin{spacing}{0.2}\end{spacing}
                \State $w_{d_k} \gets \frac{score_{d_k} \cdot norm\_score_{d_k} \cdot RS_{d_k}}{\sum_{i=1}^M score_{d_i} \cdot norm\_score_{d_k} \cdot RS_{d_i}}$
                \begin{spacing}{0.2}\end{spacing}
                \State $var \gets (W_{img} \cdot H_{img})/(W_d \cdot H_d)$
                \State $mask \gets Gauss((W_d,H_d),(X_{d_k},Y_{d_k}),var)$
                \State $D \gets D + w_{d_k} Resize(A_d \times mask, (W,H))$
            \EndFor
            \State \Return{$D$}
        \EndProcedure
    \end{algorithmic}
\end{algorithm}

\section{Experiments}\label{sec:exp}

\subsection{Datasets and Evaluation Protocols}

Extensive experiments have been conducted on the two large-scale pedestrian attribute datasets i.e., the PETA \cite{deng2014pedestrian} dataset and the RAP dataset \cite{li2016richly}. The PETA dataset includes 19,000 pedestrian samples, each annotated with 65 attributes. Following the protocol in \cite{deng2014pedestrian}, we also select 35 binary attributes for evaluation. The RAP dataset is the largest pedestrian attribute dataset so far, including 41,585 samples with 72 attributes. As implemented in \cite{li2016richly}, 51 binary attributes are used for evaluation. In test phase, images are zoomed to have fixed-size longest side without resizing or warping.

We adopt the mean accuracy (mA) as well as the example-based criteria proposed in \cite{li2016richly} as evaluation metrics. The mA is formulated as:

\begin{equation}
    \label{eq:mA}
    mA = \frac{1}{L} \sum_{i=1}^L (\frac{|TP_i|}{|P_i|} + \frac{|TN_i|}{|N_i|})
\end{equation}

where $L$ is the number of attributes, $|TP_i|$ and $|TN_i|$ are respectively the number of correctly predicted positive and negative samples of the $i-th$ attribute, and $|P_i|$ and $|N_i|$ are respectively the number of ground-truth positive and negative samples of the $i-th$ attribute.

The example-based evaluation criteria is defined as:

\begin{align}
    \label{eq:ExampleBased}
    Acc_{exam}=\frac{1}{N}\sum_{i=1}^N\frac{|Y_i \cap f(x_i)|}{|Y_i \cup f(x_i)|} \\
    Prec_{exam}=\frac{1}{N}\sum_{i=1}^N\frac{|Y_i \cap f(x_i)|}{|f(x_i)|} \\
    Rec_{exam}=\frac{1}{N}\sum_{i=1}^N\frac{|Y_i \cap f(x_i)|}{|Y_i|} \\
    F1=\frac{2 \cdot Prec_{exam} \cdot Rec_{exam}}{Prec_{exam} + Rec_{exam}}
\end{align}

where $N$ is the number of samples, $Y_i$ is the set of ground-truth positive attribute labels of the $i-th$ sample, $f(x_i)$ is the set of predicted positive attribute labels of the $i-th$ sample, and $|\cdot|$ denotes the set cardinality.

\subsection{Recognition Performance}\label{sec:AccExp}

\begin{table*}[!tbp]
    \centering
    \begin{tabular}{|*{12}{c|}}
        \hline
        \multirow{2}*{Methods}	& \multicolumn{5}{|c|}{RAP}             & \multicolumn{5}{|c|}{PETA} \\\cline{2-11}
                                & mA    & Acc   & Prec  & Rec   & F1    & mA    & Acc   & Prec  & Rec   & F1 \\
        \hline
        ELF-mm                  & 69.94 & 29.29 & 32.84 & 71.18 & 44.95 & 75.21 & 43.68 & 49.45 & 74.24 & 59.36 \\
        FC7-mm                  & 72.28 & 31.72 & 35.75 & 71.78 & 47.73 & 76.65 & 45.41 & 51.33 & 75.14 & 61.00 \\
        FC6-mm                  & 73.32 & 33.37 & 37.57 & 73.23 & 49.66 & 77.96 & 48.13 & 54.06 & 76.49 & 63.35 \\
        ACN                     & 69.66 & 62.61 & \textbf{80.12} & 72.26 & 75.98 & 81.15 & 73.66 & 84.06 & 81.26 & 82.64 \\
        DeepMAR                 & 73.79 & 62.02 & 74.92 & 76.21 & 75.56 & 82.89 & 75.07 & 83.68 & 83.14 & 83.41 \\
        DeepMAR*                & 74.44 & \textbf{63.67} & 76.53 & 77.47 & \textbf{77.00} & - & - & - & - & - \\
        \hline
        \hline
        WPAL-GoogleNet-GMP      & \textbf{81.25} & 50.30 & 57.17 & 78.39 & 66.12 & \textbf{85.50} & \textbf{76.98} & \textbf{84.07} & \textbf{85.78} & \textbf{84.90} \\
        WPAL-GoogleNet-FSPP     & 79.48 & 53.30 & 60.82 & \textbf{78.80} & 68.65 & 84.16 & 74.62 & 82.66 & 85.16 & 83.40 \\
        \hline
    \end{tabular}
    \caption{This table shows performance of 6 benchmark algorithms listed in \cite{li2016richly} and two models mentioned in this work, evaluated on RAP and PETA dataset using mA and example-based evaluation criteria. The DeepMAR* algorithm has no results on the PETA dataset because it depends on ground-truth body part annotations, which is not available on the PETA dataset.}
    \label{table:OverallAcc}
\end{table*}

For comparisons, three approaches presented in \cite{li2016richly} are used as benchmarks, including, ACN \cite{sudowe2015person}, DeepMAR \cite{li2015multi}, DeepMAR* \cite{li2016richly} and SVM with CNN features. The performance on the PETA and the RAP datasets is listed in Table~\ref{table:OverallAcc}. We can find the WPAL-network performs quite well in terms of the metric of mA, while shows some weakness as evaluated with the example-based criteria on the RAP dataset. This is because the mA criteria is less affected by false alarms on classes with fewer samples than their opposite class. Consider an attribute with too few positive samples. False positive predictions (FP) does not affect the $\frac{|TP|}{|P|}$ term in the mA formula, but affect the $|TN|$, so $\frac{|TN|}{|N|}$ decreases. However, the $|N|$ is so large that $|FP|$ becomes almost neglectable compared to it, thus the influence on total value of mA is limited. On the other hand, the example-based evaluation criteria has explicit precision terms, so the influence of false alarms are normalized. Therefore, the mA and the example-based criteria reflect different characteristics of the algorithm, and choice between them should be based on application scene. The higher mA also demonstrates that the learnt mid-level features are really effective to describe the visual characteristics of most pedestrian attributes.

\begin{figure}[!tbp]
  \centering
  \includegraphics[width=0.4\textwidth]{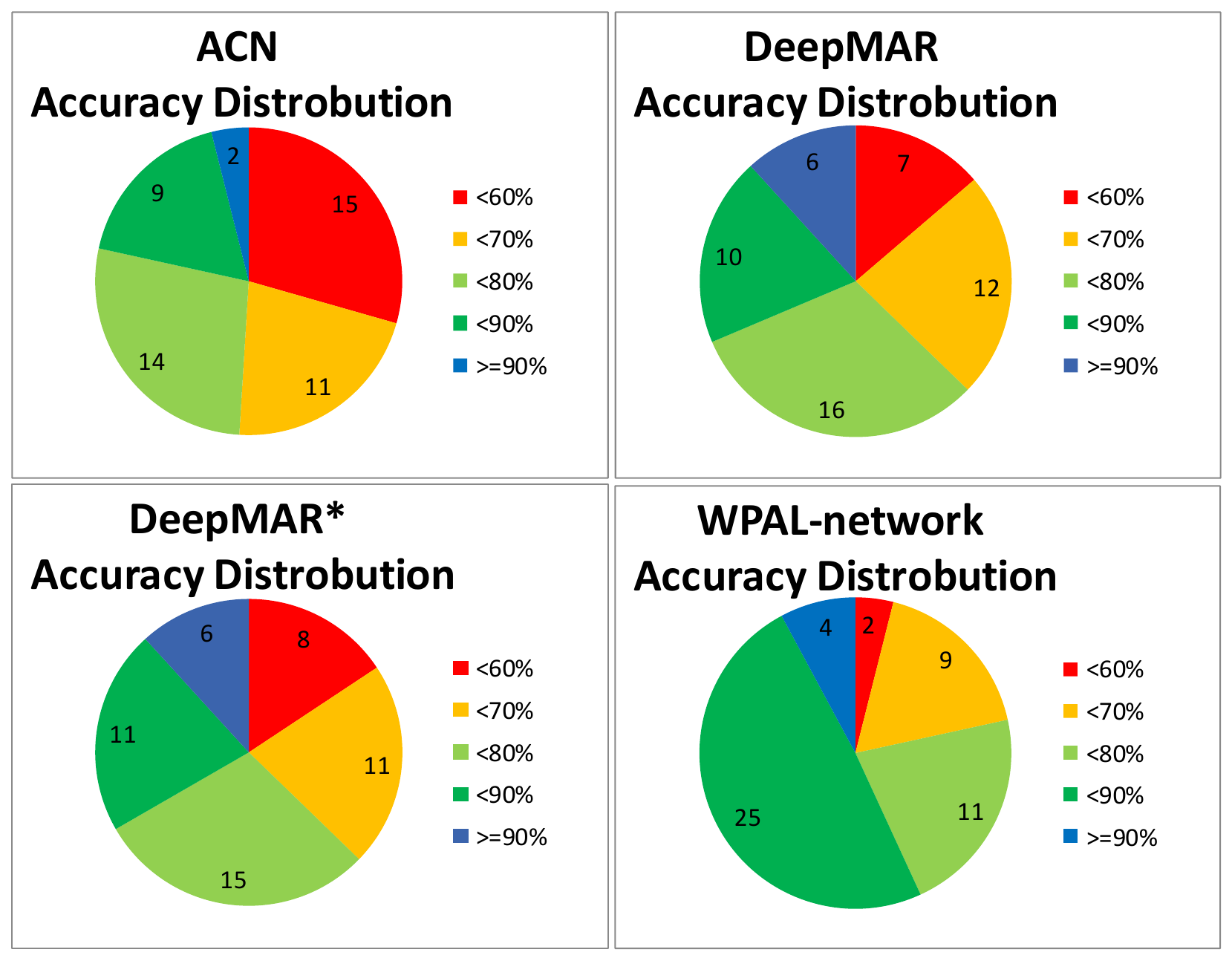}\\
  \caption{These four pie charts are the distribution summaries of independent attribute recognition accuracy of three benchmark algorithms listed in \cite{li2016richly} and the WPAL-network on the RAP dataset. The WPAL-network has more attributes recognized with accuracy between $80\%$ and $90\%$.}
  \label{fig:AccDisSum}
\end{figure}

\begin{table*}[!tbp]
    \centering
    \begin{tabular}{|*{8}{c|}}
        \hline
        Model & ELF & FC7-mm & FC6-mm & ACN & DeepMAR & DeepMAR* & WPAL \\
        \hline
		Bald Head & 71.9 & 69.6949 & 72.852 & 60.7815 & 62.3884 & 69.7718 & \textbf{84.2928} \\
		Long Hair & 78 & 79.986 & 81.4963 & 88.6562 & 90.2707 & \textbf{92.5585} & 52.5183 \\
		Hat & 69.5 & 73.3137 & 72.9159 & 57.3349 & 62.3532 & 78.0958 & \textbf{84.5464} \\
		Sweater & 59.9 & 62.2005 & 63.0389 & 58.5836 & 66.7341 & 64.8774 & \textbf{72.7116} \\
		Tights & 65.3 & 68.0768 & 69.7076 & 61.6388 & 66.4883 & 65.3833 & \textbf{76.4499} \\
		Short Skirt & 76.2 & 78.028 & 78.6267 & 70.8015 & 76.8846 & 78.0803 & \textbf{88.7883} \\
		Single-Shoulder Bag & 66.7 & 71.328 & 72.6608 & 64.7848 & 73.5757 & 71.8415 & \textbf{81.2811} \\
		Handbag & 66.4 & 71.8545 & 72.2973 & 63.1261 & 72.4594 & 68.198 & \textbf{85.2697} \\
		Box(Attachment) & 67.8 & 70.6301 & 71.4554 & 64.9486 & 72.0743 & 69.5466 & \textbf{78.651} \\
		Plastic Bag & 60.9 & 70.5147 & 70.6487 & 58.297 & 66.9855 & 61.1198 & \textbf{82.3949} \\
		Paper Bag & 63.9 & 66.7102 & 68.5074 & 53.8447 & 60.4691 & 57.7316 & \textbf{78.0773} \\
		Calling & 65.9 & 68.7387 & 70.1488 & 69.5448 & 76.8771 & 82.326 & \textbf{89.1727} \\
        \hline
    \end{tabular}
    \caption{This tables shows attributes in the RAP dataset where independent recognition accuracy differs larger than $5\%$ between our model and best of the benchmarks listed in \cite{li2016richly}. We can find that most of these attributes are better recognized by our model, but our model performs extremely bad on the Long Hair attribute.}
    \label{table:IndependentAcc}
\end{table*}

We also compare the individual attribute recognition accuracies of our model with other benchmarks. The accuracy distribution of these models is shown in Figure~\ref{fig:AccDisSum}, and Table~\ref{table:IndependentAcc} shows some selected attributes whose recognition accuracy difference between our model and the best of the benchmarks is larger than $5\%$. Recognition performance of all attributes can be found in the supplemental materials.

\subsection{Attribute Localization and Shape Estimating}

\begin{figure*}
  \centering
  \includegraphics[width=0.75\textwidth]{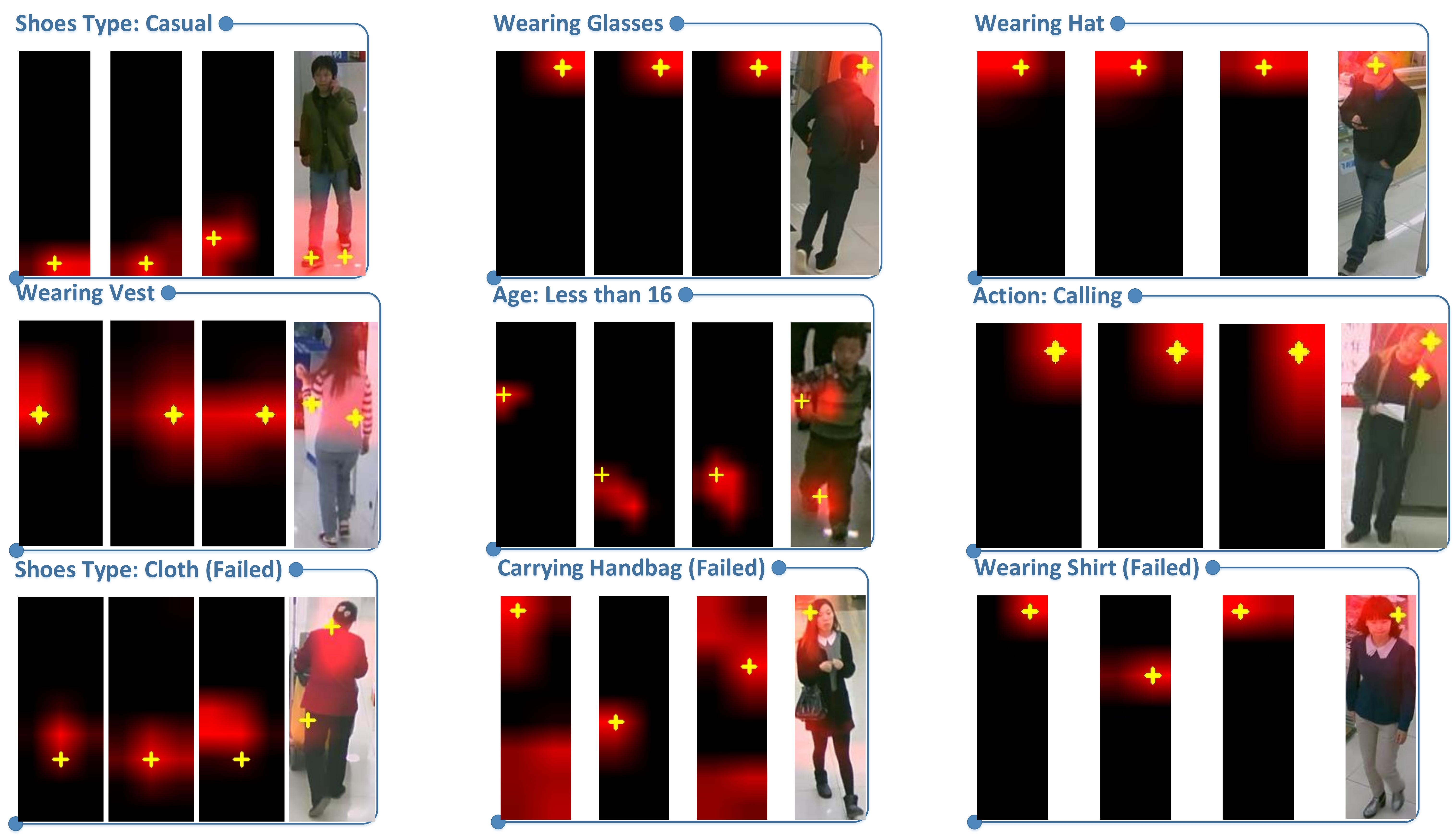}\\
  \caption{In each group, the first three columns are activation maps of selected highly-correlated detectors, and the fourth is the effect of attribute localization. The first row contains examples of accurately located fine-scale attributes. The second row contains examples of roughly located abstract or large-scale attributes. The third row contains failed cases, where the attribute labels are correctly predicted, but the locations predicted greatly differ from our expectations. Best viewed in color.}
  \label{fig:Locate}
\end{figure*}

Since there is no ground-truth data to evaluate the performance of attribute localization, we visualize some examples on pedestrian attribute localization. In Figure~\ref{fig:Locate}, the first and second rows show some successful examples, and the last row shows some failure cases. As shown in the figure, some fine-scale attributes, such as glasses, hat and shoes can be located correctly, which suggest the effectiveness of the proposed method.

\subsection{Body Shape Estimating}

\begin{figure}[!tbp]
  \centering
  \includegraphics[width=0.32\textwidth]{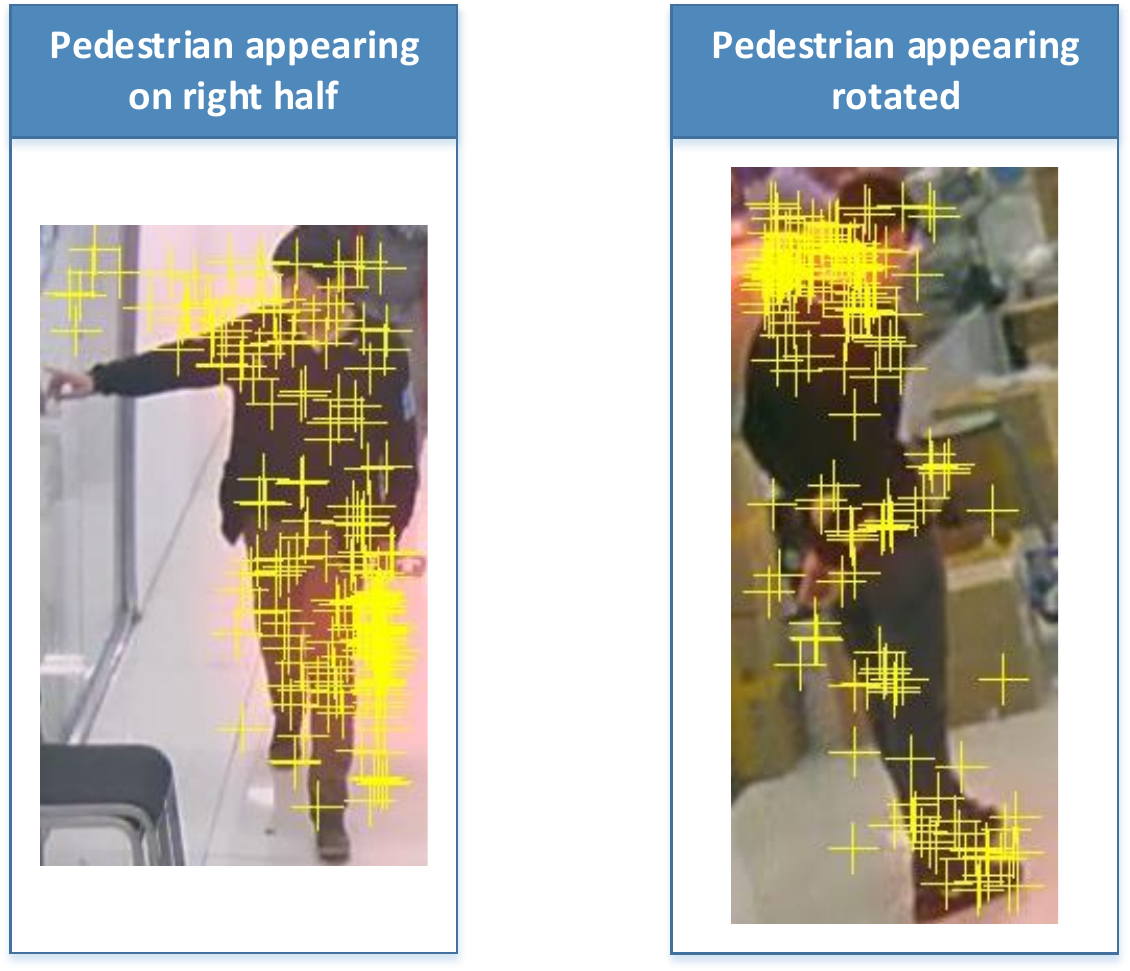}\\
  \caption{This picture shows the estimated body region of a pedestrian sample. We select two examples. In the first one, the pedestrian appears in a rotation angle. In the second sample, the pedestrian appears in the right half of the cropped image. The body trunks in abnormal status right overlap with the estimated informative region. Best viewed in color.}
  \label{fig:BodyShape}
\end{figure}

In practice, besides the pedestrian body region, the cropped image sample may contain some unrelated background contents. Sometimes, due to the large pose variation and occlusions by other environmental objects, the problem gets more serious. Thus, we expect that the recognition algorithm should have the capability to eliminate those unrelated contents by estimating a rough region of the pedestrian body. Here, based on the existence possibility maps of the learnt mid-level features, we can test this capability by estimating the informative region which contributes to the recognition of pedestrian attributes mostly. Figure \ref{fig:BodyShape} shows two examples. We can find that the informative regions in both samples overlap with the pedestrians' bodies perfectly, which illustrate that the proposed approach has the "right" attention capability to understand pedestrian attributes.

\subsection{Correlation Strength Observation}

\begin{table*}[!htbp]
    \centering
    \begin{tabular}{|*{6}{c|}}
        \hline
        \multirow{2}*{Attribute Name}	& \multicolumn{5}{|c|}{Strongly-correlated Low-level Detector Bin (Rank - Level)} \\\cline{2-6}
                                        & First                 & Second                & Third                 & Fourth                & Fifth \\
        \hline
        Action: Calling                 & $383^{th}$ - Lev.2	& $419^{th}$ - Lev.2	& $445^{th}$ - Lev.2	& $470^{th}$ - Lev.2	& $473^{th}$ - Lev.2 \\
        Has Black Hair                  & $605^{th}$ - Lev.1	& $607^{th}$ - Lev.1	& $613^{th}$ - Lev.1	& $620^{th}$ - Lev.1	& $628^{th}$ - Lev.1 \\
        Orientation: Front              & $268^{th}$ - Lev.1	& $272^{th}$ - Lev.1	& $281^{th}$ - Lev.1	& $283^{th}$ - Lev.1	& $289^{th}$ - Lev.1 \\
        Has Glasses                     & $785^{th}$ - Lev.2	& $818^{th}$ - Lev.1	& $821^{th}$ - Lev.1	& $842^{th}$ - Lev.1	& $847^{th}$ – Lev.2 \\
        Upper Part:Black                & $784^{th}$ - Lev.1	& $786^{th}$ - Lev.1	& $796^{th}$ - Lev.1	& $797^{th}$ - Lev.1	& $798^{th}$ - Lev.1 \\
        Upper Part: Red                 & $560^{th}$ - Lev.2	& $672^{th}$ - Lev.2	& $640^{th}$ - Lev.1	& $667^{th}$ - Lev.2	& $668^{th}$ - Lev.1 \\
        Lower Part: Red                 & $795^{th}$ - Lev.2	& $886^{th}$ - Lev.2	& $895^{th}$ - Lev.2	& $965^{th}$ - Lev.1	& $966^{th}$ - Lev.2 \\
        \hline
    \end{tabular}
    \caption{This table displays some selected attributes with top-5 highly ranking low-level detector bins correlated to them. Each item in the table is a tuple of the bin's correlation ranking and the level of the detector it belongs to. Note that there are totally 14336 detector bins in the WPAL-network, with 5120 bins on level 1 and 2 respectively and 4096 bins on level 3.}
    \label{table:CorrObservation}
\end{table*}

Although we expect low-level detectors to help recognition of low-level attributes, by observing the matrix of correlation strengths between attributes and mid-level feature detectors, we find that high-level detectors still play the most significant role in prediction, no matter for high-level attributes or low-level attributes. However, there are still some low-level detector bins ranking relatively high in the sorted bin list of some attributes by correlation strength to them, meaning that low-level features do make strong contribution to the recognition of attributes. Table~\ref{table:CorrObservation} shows some selected features with top-5 highly ranking low-level detector bins correlated to them.

\section{Conclusion and Future Work}

In this work, we formulate the pedestrian attribute recognition into a weakly supervised object detection framework at the first time. A novel WPAL-network is proposed for the tasks of pedestrian attribute recognition and localization. Instead of directly predicting multiple attributes, we firstly discover a set of mid-level attribute-relevant features, and then predict attributes based of the response of these features. Furthermore, the activation maps of these features can be used to infer the location and rough shape of an attribute. The competitive recognition performance on the two large-scale attribute datasets demonstrates the effectiveness of the proposed WPAL-network.

In the future, we will seek more powerful detectors utilizing additional information such as background context and location relationship between discovered mid-level features to improve accuracy and solve recognition failure on attributes like long hair.

{\small
\bibliographystyle{ieee}
\bibliography{egbib}
}

\end{document}